\documentclass[a4paper,onecolumn]{article}

\usepackage{graphicx}
\usepackage{amssymb,amsmath,bm}
\usepackage{textcomp}
\usepackage{enumitem}
\usepackage[margin=1.2in]{geometry}
\usepackage{authblk}
\DeclareMathOperator{\trace}{tr}

\setlist[itemize]{leftmargin=*}

\def\vec#1{\ensuremath{\bm{{#1}}}}

\newcommand{\si}{{s_i}}
\newcommand{\ci}{{c_i}}

\DeclareMathOperator{\LR}{{LR}}
\DeclareMathOperator{\LLR}{{LLR}}
\DeclareMathOperator{\sumexp}{{sumexp}}

\title{Joint Probabilistic Linear Discriminant Analysis}

\author[1]{Luciana Ferrer}
\affil[1]{\mbox{Instituto de Investigaci\'on en Ciencias de la Computaci\'on, }
\mbox{CONICET-UBA, Argentina}}
\date{}
\begin{document}

  \maketitle

\begin{abstract}
Standard probabilistic linear discriminant analysis (PLDA) for speaker recognition assumes that the sample's features (usually, i-vectors) are given by a sum of three terms: a term that depends on the speaker identity, a term that models the within-speaker variability and is assumed independent across samples, and a final term that models any remaining variability and is also independent across samples. In this work, we propose a generalization of this model where the within-speaker variability is not necessarily assumed independent across samples but dependent on another discrete variable. This variable, which we call the channel variable as in the standard PLDA approach, could be, for example, a discrete category for the channel characteristics, the language spoken by the speaker, the type of speech in the sample (conversational, monologue, read), etc. The value of this variable is assumed to be known during training but not during testing. Scoring is performed, as in standard PLDA, by computing a likelihood ratio between the null hypothesis that the two sides of a trial belong to the same speaker versus the alternative hypothesis that the two sides belong to different speakers. The two likelihoods are computed by marginalizing over two hypothesis about the channels in both sides of a trial: that they are the same and that they are different. This way, we expect that the new model will be better at coping with same-channel versus different-channel trials than standard PLDA, since knowledge about the channel (or language, or speech style) is used during training and implicitly considered during scoring. 
\end{abstract}

\section{Introduction}

PLDA \cite{prince:plda} was first proposed for doing inferences about the identity of a person from an image of their face. The technique was later widely adopted by the speaker recognition community, becoming the state-of-the-art scoring technique for this task. In this work, we will adopt the nomenclature usually used in the speaker recognition community. Yet, the model proposed can be used for the original image processing task or any other task for which standard PLDA is used.

Standard PLDA assumes that the vector $m_i$ representing a certain sample (in speaker recognition these will usually be i-vectors \cite{Dehak11}) from speaker $s_i$ is given by
\begin{equation}
m_i = V y_\si + U x_i + z_i
\end{equation}
where $y_\si$ is a vector of size $R_y$ (the dimension of the speaker subspace) and $x_i$ is a vector of size $R_x$ (the dimension of the channel subspace), and
\begin{eqnarray}
y_\si & \sim & N(0,I) \\
x_i & \sim & N(0,I) \\
z_i & \sim & N(0, D^{-1})
\end{eqnarray}
where the matrix $D$ is assumed to be diagonal. All these latent variables are assumed independent: speaker variables are independent across speakers and the channel variable $x_i$ and noise variable $z_i$ are independent across samples. 

The model described corresponds to the original PLDA formulation. In speaker recognition a simplified version of PLDA is more commonly used, where the matrix $V$ is full rank and the channel factor is absorbed into the noise factor, which is then assumed to have a full rather than diagonal covariance matrix. This simpler model has shown to give better performance than the original model. See \cite{sizov2014} for a comprehensive explanation of the usual flavors of PLDA. 

In this work, we propose a generalization of the original model where the channel variable is no longer considered independent across samples, but potentially shared (tied) across samples from different speakers. 

\section{Proposed Joint PLDA Model}

The proposed generalization of the PLDA model implies that the channel latent variable is no longer dependent only on the sample index, but rather, depends on a separate channel label. This makes the model symmetric in the two latent variables (speaker and channel, in our nomenclature) in the sense that both variables are tied across all samples sharing a certain label. To represent this dependency, we introduce a channel label for each sample, called $\ci$. Given this channel label, and the speaker label $\si$, we propose to model vector $m_i$ for sample $i$ as:
\begin{equation}
m_i = V y_\si + U x_\ci + z_i
\end{equation}
where, as before, $y_\si$ is a vector of size $R_y$ and $x_\ci$ is a vector of size $R_x$, and
\begin{eqnarray}
y_\si & \sim & N(0,I) \\
x_\ci & \sim & N(0,I) \\
z_i & \sim & N(0, D^{-1})
\end{eqnarray}
The model's parameters to estimate are $\lambda = \{V, U, D\}$, as in the standard PLDA formulation, but the input data for the training algorithm is now expected to have a second set of labels indicating the channel identity of each sample. Note that while we call this variable the channel, for consistency with previous work on PLDA, the channel could be anything that is a nuisance variable for the task. For speaker recognition this could be, for example, a discrete classification of the channel type itself, or the language spoken, the speech style, etc. 

The following sections derive the probabilities that are needed during training with the expectation-maximization algorithm and during scoring with the likelihood ratio. The derivations closely follow the ones for standard PLDA in \cite{em4plda} with one main difference: in the new model, most probabilities cannot be formulated by speaker and then multiplied to get the total probabilities, as is usually done for standard PLDA, since the channel introduces dependencies across samples from different speakers. Instead, we formulate all probabilities over all samples.

In the following, we take
\begin{eqnarray}
Y & = & \{y_1, \ldots, y_S \} \\
X & = & \{x_1, \ldots, x_C \} 
\end{eqnarray}
where $S$ is the total number of speakers and $C$ is the total number of channels. 

\subsection{Prior}

The joint prior for the hidden variables for all the data is given by
\begin{equation}
\label{eq:prior}
p(Y,X) = p(X) p(Y) \propto \exp(-\frac{1}{2} \sum_s y^T_s y_s -\frac{1}{2} \sum_c x^T_c x_c) 
\end{equation}

\subsection{Likelihood}

The full data likelihood is given by

\begin{eqnarray}
\label{eq:llk} 
p(M|Y,X,\lambda) & = & \prod_i N(m_i|Vy_\si+Ux_\ci,D^{-1}) \nonumber \\
& \propto &   \exp  \sum_i \left(-\frac{1}{2} (m_i - Vy_\si -Ux_\ci)^T D (m_i - Vy_\si -Ux_\ci) + \frac{1}{2} \log|D| \right) \nonumber \\
 & = & \exp \sum_i \left( -\frac{1}{2} m^T_i D m_i + m^T_i D V y_\si + m^T_i D U x_\ci \right. \nonumber \\
 & & \left. -\frac{1}{2}  y^T_\si V^T D V y_\si - y^T_\si V^T D U x_\ci  -\frac{1}{2}  x^T_\ci U^T D U x_\ci  + \frac{1}{2} \log|D|         \right) 
\end{eqnarray}

\subsection{Joint}

The joint probability is given by the product of the likelihood and the prior,
\begin{eqnarray}
p(M,Y,X|\lambda) & \propto & \exp \left[ \sum_i \left( -\frac{1}{2} m^T_i D m_i + m^T_i D V y_s + m^T_i D U x_c \right. \right. \nonumber\\
 & & \left. -\frac{1}{2}  y^T_\si V^T D V y_\si - y^T_\si V^T D U x_\ci  -\frac{1}{2}  x^T_\ci U^T D U x_\ci  \right) \nonumber\\
 & & \left. -\frac{1}{2} \sum_s y^T_s y_s -\frac{1}{2} \sum_c x^T_c x_c \right] \nonumber \\
 & = & \exp \left[ \sum_i \left( -\frac{1}{2} m^T_i D m_i + m^T_i D V y_\si + m^T_i D U x_\ci - x^T_\ci J y_\si \right) \right. \nonumber\\
 & & \left. -\frac{1}{2} \sum_s y^T_s L_s y_s -\frac{1}{2} \sum_c x^T_c K_c x_c \right] 
\end{eqnarray}
where 
\begin{eqnarray}
J & = & U^T D V \\
K_c & = & n_c U^T D U + I \label{eq:precx}\\
L_s & = & n_s V^T D V + I \label{eq:precy}
\end{eqnarray}
where $n_c$ is the number of samples for channel $c$ and $n_s$ is the number of samples for speaker $s$.

\subsection{Posterior}

We compute the posterior from two factors:
\begin{equation}
p(Y,X|M,\lambda) = p(Y|X,M,\lambda) p(X|M,\lambda)
\end{equation}

\subsubsection{Outer posterior}
\label{sec:posteriors}

The outer posterior is proportional (as a function of $Y$) to the joint probability. Keeping only the terms in the joint probability that depend on $Y$ we get
\begin{eqnarray}
p(Y|X,M,\lambda) & \propto & p(M,X,Y|\lambda) \nonumber \\
	& \propto & \exp \left[ \sum_i \left(  m^T_i D V y_\si - x^T_\ci J y_\si \right)  -\frac{1}{2} \sum_s y^T_s L_s y_s \right] \nonumber\\
	& \propto & \exp \left[ \sum_i (m^T_i D V - x^T_\ci J) y_\si  -\frac{1}{2} \sum_s y^T_s L_s y_s \right] \nonumber\\
	& \propto & \exp \sum_s \left[  y^T_s \sum_{i|\si=s} (V^T  D m_i - J^T x_\ci)  -\frac{1}{2} y^T_s L_s y_s \right] \nonumber\\
	& \propto & \exp \sum_s \left[  y^T_s (V^T  D f_s - J^T \bar{x}_s)  -\frac{1}{2} y^T_s L_s y_s \right] \nonumber\\
	& \propto & \prod_s N(y_s | \hat{y}_s, L_s^{-1})
\end{eqnarray}
where 
\begin{eqnarray}
f_s & = &\sum_{i|\si=s} m_i \\
\bar{x}_s & = & \sum_{i|\si=s}  x_\ci \\
\tilde{y}_s & = & L_s^{-1} V^T  D f_s  \\
\hat{y}_s & = & \tilde{y}_s - L_s^{-1} J^T \bar{x}_s \label{eq:meany}
\end{eqnarray}

\subsubsection{Inner posterior}

To get the inner posterior, which is proportional to the joint between $X$ and $M$, we use a nice trick \cite{besag1989}:
\begin{eqnarray}
p(X|M,\lambda) & \propto & p(M,X|\lambda)  = \left. \frac{p(Y,X,M|\lambda)}{p(Y|X,M,\lambda)}\right|_{Y=0} \nonumber \\
& \propto & \frac{\exp \left(\sum_i m^T_i D U x_\ci -\frac{1}{2} \sum_c x^T_c K_c x_c \right)}{\exp\left( -\frac{1}{2} \sum_s \hat{y}_s^T L_s \hat{y}_s\right)}
\end{eqnarray}
where the right hand side in the first line is independent of $Y$ (since the left-hand side is) and, hence, can be conveniently evaluated at 0. Now, we expand the exponent in the denominator 
\begin{eqnarray}
\sum_s \hat{y}_s^T L_s \hat{y}_s & = & \sum_s  (\tilde{y}_s^T -  \bar x^T_s J L_s^{-1}) (L_s \tilde{y}_s -  J^T \bar x_s ) \nonumber\\
& = &  \sum_s  (-  2 \bar{x}_s^T  J \tilde{y}_s + \bar{x}_s^T J L_s^{-1} J^T \bar{x}_s ) + \text{const}
\end{eqnarray}
and plug it back in the posterior:
\begin{eqnarray}
p(X|M,\lambda) & \propto & \exp \left(\sum_i m^T_i D U x_\ci -\frac{1}{2} \sum_c x^T_c K_c x_c - \sum_s  \bar{x}_s^T  J \tilde{y}_s +  \frac{1}{2} \sum_s  \bar{x}_s^T J L_s^{-1} J^T \bar{x}_s    \right) \nonumber\\
 & = & \exp \left(\sum_c g_c^T D U x_c -\frac{1}{2} \sum_c x^T_c K_c x_c - \sum_i  x_\ci^T  J \tilde{y}_\si +  \frac{1}{2}  \sum_s \bar{x}_s^T J L_s^{-1} J^T \bar{x}_s    \right) \nonumber\\
 & = & \exp \left(\sum_c g_c^T D U x_c -\frac{1}{2} \sum_c x^T_c K_c x_c - \sum_c  x_c^T  J \bar{\tilde{y}}_c +  \frac{1}{2}  \sum_s \bar{x}_s^T J L_s^{-1} J^T \bar{x}_s    \right) 
 \end{eqnarray}
where 
\begin{eqnarray}
g_c & = &\sum_{i|\ci=c} m_i \\
\bar{\tilde{y}}_c & = & \sum_{i|\ci=c}  \tilde{y}_\si =  \sum_{i|\ci=c}  L_\si^{-1} V^T  D f_\si
\end{eqnarray}

Since there is no way to disentangle the $x$s for different $c$s, we need to obtain the distribution for all $x$s at once. We define vectors which are the concatenation of all individual vectors:
\begin{eqnarray}
\vec X & = & [x_1^T \ldots x_C^T]^T \\
\vec Y & = & [y_1^T \ldots y_S^T]^T 
\end{eqnarray}
and, similarly, for all other vectors. Converting the sums into matrix form, we get
\begin{eqnarray}
p(X|M,\lambda) & \propto & \exp \left(\vec X^T M_1 \vec G -\frac{1}{2} \vec X M_2 \vec X - \vec X  M_3 \bar{\tilde{\vec Y}} +  \frac{1}{2}  \bar{\vec X}^T M_4 \bar{\vec X}    \right) \nonumber\\
 & = & \exp \left(\vec X^T M_1 \vec G -\frac{1}{2} \vec X M_2 \vec X - \vec X  M_3 \bar{\tilde{\vec Y}} +  \frac{1}{2}  {\vec X}^T H^T M_4  H \vec X \right) \nonumber\\
 & = & \exp \left(\vec X^T \Phi  -\frac{1}{2} \vec X (M_2 - H^T M_4  H ) \vec X  \right) 
 \end{eqnarray}
where 
\begin{eqnarray}
M_1 & = & \text{diagn}(U^TD, C) \\
M_2 & = & \text{diag}(K_1, \ldots, K_C) \\
M_3 & = & \text{diagn}(J, C) \label{eq:m3} \\
M_4 & = & \text{diag}( J L_1^{-1} J^T , \ldots, J L_S^{-1} J^T) \label{eq:m4} \\
\Phi & = & M_1 \vec G - M_3 \bar{\tilde{\vec Y}} \label{eq:Phi} \\
\bar{\vec X} & = & H \vec X \label{eq:vecxbar}
\end{eqnarray}
where $\text{diagn}(M,N)$ is a block diagonal matrix with matrix $M$ in each of $N$ blocks and $\text{diag}(M_1, \ldots, M_N)$ is a block diagonal matrix with blocks given by matrices $M_i$. The matrix H is of size $SR_x$x$CR_x$, where block $H_{s,c}$ ($R_x$ rows and columns starting at position $(sR_x,cR_x)$ in $H$) is given by:
\begin{equation}
H_{s,c} = n_{s,c} I
\end{equation}
where $I$ is the identity matrix of size $R_x$ and $n_{s,c}$ is the number of times that channel $s$ occurs for speaker $s$, which could be zero.

Hence, finally:

\begin{eqnarray}
\label{eq:innerpost}
p(\vec X|M,\lambda) & = & N(\vec X | \hat{\vec X}, \Sigma) 
 \end{eqnarray}
where
\begin{eqnarray}
\Sigma & = & (M_2 - H^T M_4  H)^{-1} \label{eq:Sigma} \\
\hat{\vec X} & = & \Sigma  \Phi
 \end{eqnarray}

Now if we need the distribution of $x_c$ we just need to get the marginal distribution from the one above, which means getting the $c$ block from the mean and covariance matrix. So,

\begin{eqnarray}
p(x_c|M,\lambda) & = & N(x_c | \hat{x}_c, \Sigma_c)
 \end{eqnarray}

\begin{eqnarray}
\Sigma_c & = & \text{block}(\Sigma,c,c) \label{eq:sigmax}\\
\hat{x}_c & = &  \text{block}(\hat{\vec X} ,c) \label{eq:meanx} 
 \end{eqnarray}

\section{EM algorithm}

As in standard PLDA, we will use the expectation-maximization algorithm to train the model parameters. 

\subsection{EM objective}

The objective function of EM is the likelihood of the data given the model, which can be obtained as follows:
\begin{eqnarray}
\log p(M|\lambda) & = & \log \left. \frac{p(M|Y,X,\lambda)p(X)p(Y)}{p(Y|X,M,\lambda)p(X|M,\lambda)}\right|_{Y=0,X=0} \nonumber \\
& = & \sum_i \left( -\frac{1}{2} m^T_i D m_i + \frac{1}{2} \log|D| \right) - \sum_s \left( -\frac{1}{2} \tilde y^T_s L_s \tilde y_s + \frac{1}{2} \log|L_s| \right) \nonumber \\
& & + \frac{1}{2} \hat{\vec{X}}^T \Sigma^{-1} \hat{\vec X} + \frac{1}{2} \log|\Sigma| + \mbox{constant} 
\end{eqnarray}
Terms that depend only on data or are constant have been discarded since they will not change with model's parameters.

\subsection{EM auxiliary function}
The EM auxiliary function is given by the expected value of the log-likelihood with respect to the posterior probability of the  hidden variables given the data and the previously estimated model parameters, $\lambda_{k-1}$. Defining $Z=\{X,Y\}$, then
\begin{eqnarray}
Q(\lambda_k|\lambda_{k-1}) & = & E_{Z|M,\lambda_{k-1}} \left[\log p(M,Z|\lambda_k)\right] \nonumber\\
& = & \langle \log p(M|Z,\lambda_k) p(Z|\lambda_k) \rangle \nonumber \\
& = & \langle \log p(M|Z,\lambda_k) \rangle + \text{const} \nonumber \\
& = & \sum_i  \langle \frac{1}{2} \log|D| -  \frac{1}{2} (m_i - W z_i)^T D (m_i - W z_i)\rangle + \text{const} \nonumber \\
& = & \sum_i  \langle   \frac{1}{2} \log|D| -  \frac{1}{2} m_i^T D m_i -  \frac{1}{2} z_i^T W^T D W z_i + m_i^T D W z_i \rangle + \text{const} \nonumber\\
& = &  \frac{N}{2} \log|D| -  \frac{1}{2} \sum_i \langle m_i^T D m_i -  \frac{1}{2} z_i^T W^T D W z_i + m_i^T D W z_i \rangle + \text{const} \nonumber\\
& = &  \frac{N}{2} \log|D| -  \frac{1}{2} \trace(S D) -  \frac{1}{2} \trace(R W^T D W) + \trace(T D W)  + \text{const} 
\end{eqnarray}
where the $\langle$ and $\rangle$ symbols stand for the expectation with respect to the distribution of $Z$ given the data and the previous parameters, as in the first line, and
\begin{eqnarray}
z_i &=& [x_\ci^T y_\si^T]^T \\
W & = & [U V] \\
S & = & \sum_i m_i m_i^T \\
R & = & \sum_i \langle z_i z_i^T \rangle \\
T & = & \sum_i \langle z_i \rangle m_i^T
\end{eqnarray}
In this derivation we use the fact that $p(Z|\lambda_k)$ is a constant with respect to $\lambda_k$, since the prior for the latent variables does not depend on model's parameters (Equation (\ref{eq:prior})).
We also use the fact that $m_i^T D m_i$ is a scalar and, hence, $m_i^T D m_i = \trace(m_i^T D m_i) = \trace(m_i m_i^T D)$ since $\trace(AB) = \trace(BA)$. A similar thing is done for the $R$ term and the T term.

\subsection{M-Step}

Now, differentiating $Q$ with respect to $D$ and $W$ and setting to zero, we get that
\begin{eqnarray}
D^{-1} & = & \frac{1}{N} (S - WT) \\
W^T & = & R^{-1} T
\end{eqnarray}
So, the matrices are estimated exactly the same way as in the standard PLDA approach \cite{em4plda}. The complexity lies in getting $R$ and $T$.

\subsection{E-Step}

To find $T$ and $R$ we decompose it into their $x$ and $y$ blocks and then use the distributions we found in Section \ref{sec:posteriors}. 

For the $x$ blocks we have: 
\begin{eqnarray}
\langle x_c \rangle &= &\hat{x}_c,\\
\langle x_c x_c^T \rangle &= &\Sigma_c + \hat x_c \hat x_c^T, 
\end{eqnarray}
where $\hat{x}_c$ is given by Equation (\ref{eq:meanx}), and $\Sigma_c$ is given by Equation (\ref{eq:sigmax}). 

For the expectation of $y$ we use the law of total expectation:
\begin{eqnarray}
\langle y_s \rangle & = & E_{Y|M,\lambda_{k-1}}\left[y_s\right] \nonumber \\
& = & E_{X|M,\lambda_{k-1}}\left[ E_{Y|M,X,\lambda_{k-1}}\left[y_s|X\right]\right] \nonumber \\
& = & E_{X|M,\lambda_{k-1}}\left[ L_s^{-1} V^T  D f_s - L_s^{-1} J^T \bar{x}_s\right] \nonumber \\
& = & L_s^{-1} V^T  D f_s - L_s^{-1} J^T E_{X|M,\lambda_{k-1}}\left[\sum_{i|\si=s}  x_\ci\right] \nonumber \\
& = & L_s^{-1} V^T  D f_s - L_s^{-1} J^T \sum_{i|\si=s}  \hat{x}_\ci
\end{eqnarray}
where we use Equation (\ref{eq:meany}) to get the inner expectation in the second line. Similar procedures are used to get the second moments needed to compute $R$ later in this section.

Using all the equalities above, we can compute the two components of $T$:
\begin{eqnarray}
T_x & = & \sum_i \langle x_\ci \rangle m_i^T \nonumber\\
&=& \sum_c \hat x_c \sum_{i|\ci=c} m_i^T \nonumber\\
&=& \sum_c \hat x_c g_c^T \\
T_y & = & \sum_i \langle y_\si \rangle m_i^T \nonumber\\
&=& \sum_s \hat y_s \sum_{i|\si=s} m_i^T \nonumber\\
& = & \sum_s (L_s^{-1} V^T  D f_s - L_s^{-1} J^T \sum_{i|\si=s}  \hat x_\ci ) f_s^T \nonumber\\
& = &  \sum_s  L_s^{-1} V^T  D f_s f_s^T - \sum_s L_s^{-1} J^T \sum_{i|\si=s}  \hat x_\ci  f_s^T \nonumber\\
& = &  \sum_s  L_s^{-1} V^T  D f_s f_s^T - \sum_s L_s^{-1} J^T \bar{\hat x}_s  f_s^T 
\end{eqnarray}
where we define
\begin{eqnarray}
\bar{\hat x}_s  = \sum_{i|\si=s}  \hat x_\ci 
\end{eqnarray}
Finally, we can get the components of $R$ as follows:
\begin{eqnarray}
R_{xx} & = & \sum_i \langle x_\ci x_\ci^T\rangle \nonumber \\
& = & \sum_c n_c \langle x_c x_c^T\rangle \nonumber\\
&=& \sum_c n_c (\Sigma_c + \hat x_c \hat x_c^T)\\
\nonumber\\
R_{yx} & = & \sum_i \langle y_\si x_\ci^T\rangle \nonumber\\
&=& \sum_i \langle ( L_\si^{-1} V^T  D f_\si - L_\si^{-1} J^T \sum_{j|s_j=\si}  x_{c_j} ) x_\ci^T \rangle \nonumber\\
&=& \sum_i L_\si^{-1} \left[ V^T  D f_\si \hat x_\ci^T- J^T \sum_{j|s_j=\si}  \langle x_{c_j} x_\ci^T \rangle \right] \nonumber\\
&=& \sum_s L_s^{-1} \left[ V^T  D f_s \sum_{i|\si=s}  \hat x_\ci^T- J^T  \sum_{i|\si=s} \sum_{j|s_j=s}  \langle x_{c_i}  x_{c_j}^T \rangle \right] \nonumber\\
&=& \sum_s L_s^{-1} \left[ V^T  D f_s \bar{\hat x}_s^T- J^T  \langle \bar x_s  \bar x_s^T \rangle \right]\\
\nonumber\\
R_{yy} & = & \sum_i \langle y_\si y_\si^T\rangle \nonumber\\
& = & \sum_s n_s \langle  y_s y_s^T\rangle \nonumber\\
& = & \sum_s n_s \left[L_s^{-1} +\langle \hat y_s(\vec X) \hat y_s^T(\vec X)\rangle\right] 
\end{eqnarray}

where 

\begin{eqnarray}
\langle \hat y_s(\vec X) \hat y_s^T(\vec X)\rangle  & = & E[L_s^{-1} (V^T D f_s - J^T \bar x_s) (f_s^T D V - \bar x_s^T J) L_s^{-1}] \nonumber\\
 & = & L_s^{-1} (V^T D f_s f_s^T D V - V^T D f_s \bar{\hat x}_s^T J - J \bar{\hat x}_s f_s^T D V  + J^T \langle \bar x_s \bar x_s^T \rangle J) L_s^{-1} 
\end{eqnarray}

and 

\begin{eqnarray}
\langle \bar x_s \bar x_s^T \rangle & = & \sum_{i|\si=s} \sum_{j|s_j=s}  \langle x_{c_j} x_\ci^T \rangle \nonumber\\
& = & \sum_{i|\si=s} \sum_{j|s_j=s} [ \hat x_{c_i} \hat x_{c_j}^T + \text{block}(\Sigma,c_j,c_i)]
\end{eqnarray}

\section{Scoring}

In scoring, given two sets of i-vectors $E$ and $T$ for enrollment and test, we need to compute:
\begin{eqnarray}
\LR & = & \frac{p(E,T|H_{SS})}{p(E,T | H_{DS})}
\end{eqnarray}
where $SS$ stands for same speaker and $DS$ for different speaker. 

\subsection{Scoring for single-enrollment and single-test trials}

In standard PLDA, the denominator can be factorized as $p(E)p(T)$ but this is only because it is assumed that channel factors are independent across samples. Since we do not assume that, we need to compute LR as follows, where we assume $E$ and $T$ are single i-vectors rather than sets.
\begin{eqnarray}
\LR & = & \frac{p(E,T|H_{SS},H_{SC})P(H_{SC}|H_{SS}) + p(E,T|H_{SS},H_{DC})P(H_{DC}|H_{SS})}{p(E,T|H_{DS},H_{SC})P(H_{SC}|H_{DS}) + p(E,T|H_{DS},H_{DC})P(H_{DC}|H_{DS})}
\end{eqnarray}
Here, $SC$ stands for same channel and $DC$ for different channel. The priors for these two hypothesis given the $DS$ and $SS$ hypothesis would be task-dependent (i.e., similarly to the same-gender and different-gender priors for gender-based mixture PLDA \cite{senoussaoui2011}). 

Note that the formulation would become more complex if there was more than one sample allowed in testing or enrollment, since there could be any combination of channels for those samples, some of them being the same, some different. For now, we will focus on the case in which both $E$ and $T$ are single i-vectors $\{m_E\}$ and $\{m_T\}$. 

Define $M=\{E,T\}$. We can now write:
\begin{eqnarray}
\label{eq:likM}
p(M|H_{*S},H_{*C}) = \left. \frac{p(M|X_{H_{*C}}, Y_{H_{*S}}) p(X_{H_{*C}})p(Y_{H_{*S}})}{p(X_{H_{*C}}, Y_{H_{*S}} | M) } \right|_{X_{H_{*C}}=0, Y_{H_{*S}}=0}
\end{eqnarray}
where we use ``*'' to indicate either ``S'' (same) or ``D'' (different). On the right hand side, we drop the conditioning to the hypotheses to simplify notation. The latent variables are given by:
\begin{equation}
X_{H_{*C}} = \begin{cases} 
\{x\},             & \mbox{if } H_{*C} = H_{SC} \\
\{x_E, x_T\}, & \mbox{if } H_{*C} = H_{DC}
\end{cases}
\end{equation}
and
\begin{equation}
Y_{H_{*C}} = \begin{cases} 
\{y\},             & \mbox{if } H_{*S} = H_{SS} \\
\{y_E, y_T\}, & \mbox{if } H_{*S} = H_{DS}
\end{cases}
\end{equation}
Now, the likelihood in the numerator of Equation \ref{eq:likM} is the same for all four combination of hypotheses since, regardless of whether the latent variables are tied or not, Equation (\ref{eq:llk}) has the same form. Hence, that term cancels out in the computation of the LR. The priors, on the other hand, will have one factor for the same-speaker or same-channel case and two identical factors, once evaluated at 0, for the different-speaker or different-channel case. So,
\begin{eqnarray}
\LR & = & \left. \frac{p(x)p(y)p(X_{H_{SC}}, Y_{H_{SS}}|M)^{-1}P_{S,S}+ p(x)^2p(y)p(X_{H_{DC}}, Y_{H_{SS}}|M)^{-1}P_{D,S}}
                      {p(x)p(y)^2p(X_{H_{SC}}, Y_{H_{DS}}|M)^{-1}P_{S,D}+ p(x)^2p(y)^2p(X_{H_{DC}}, Y_{H_{DS}}|M)^{-1}P_{D,D}} \right|_{0} \nonumber \\
\end{eqnarray}
were we have shortened the names for the hypothesis priors: $P_{S,S} = P(H_{SC}|H_{SS})$, $P_{D,S} = P(H_{DC}|H_{SS})$, and so on. The evaluation at zero is done for all $X$s and $Y$s. 

All that is left to do is compute the posteriors, which are given by the product of the inner and outer posteriors, and evaluate them at 0. 
\begin{eqnarray}
p(X_{H_{*C}}, Y_{H_{*S}}|M) = p(Y_{H_{*S}}|X_{H_{*C}},M) p(X_{H_{*C}}|M)
\end{eqnarray}
The outer posterior $p(Y_{H_{*S}}|X_{H_{*C}},M)$ evaluated at 0 will be the same regardless of whether the channels are the same or different.
\begin{equation}
\left. p(Y_{H_{*S}}|X_{H_{*C}},M)\right|_0  = \begin{cases} 
N(y=0|L_S^{-1}V^TD(m_E+m_T),L_S^{-1}), & \mbox{if } H_{*S} = H_{SS} \\
N(y=0|L_D^{-1}V^TDm_E,L_D^{-1})N(y=0|L_D^{-1}V^TDm_T,L_D^{-1}), & \mbox{if } H_{*S} = H_{DS}
\end{cases}
\end{equation}
where $L_D = V^TDV+I$ and $L_S = 2  V^TDV+I$. Now we define
\begin{eqnarray}
\tilde m_E & = & V^TD m_E \nonumber\\
\tilde m_T & = & V^TD m_T \nonumber\\
k & = & \log(2\pi) \nonumber
\end{eqnarray}
and rewrite the outer posterior as
\begin{equation}
\left. \log p(Y_{H_{*S}}|X_{H_{*C}},M)\right|_0  = \begin{cases} 
-\frac{1}{2}R_y k + \frac{1}{2}\log |L_S| - \frac{1}{2} (\tilde m_E + \tilde m_T)^T L_S^{-1}  (\tilde m_E + \tilde m_T), & \mbox{if } H_{*S} = H_{SS} \\
-R_yk + \log |L_D| - \frac{1}{2} \tilde m_E^T L_D^{-1}  \tilde m_E- \frac{1}{2} \tilde m_T^T L_D^{-1}  \tilde m_T , & \mbox{if } H_{*S} = H_{DS} \\
\end{cases}
\end{equation}

The inner posterior is given in Equation (\ref{eq:innerpost}), with $\vec X = x$ for the same-channel case and $\vec X = [x_E^T x_T^T]^T$ for the different-channel case. These posteriors also depend on whether the speakers are the same or not (through the implicit conditioning on the hypothesis that we dropped in (\ref{eq:likM})), which comes into play when evaluating Equations (\ref{eq:m3}), (\ref{eq:m4}) and (\ref{eq:vecxbar}). The vectors and matrices required to compute (\ref{eq:innerpost}) are given by
\begin{eqnarray}
\vec G & = & \begin{cases}
m_E+m_T, & \mbox{if } n_c=1 \\
[m_E^T m_T^T]^T, & \mbox{if } n_c=2 \\
\end{cases} \nonumber \\
\bar{\tilde{\vec Y}} & =  &\begin{cases}
2 L_1^{-1}V^TD(m_E+m_T), & \mbox{if } n_c=1, n_s=1 \\
L_2^{-1}V^TD(m_E+m_T), & \mbox{if } n_c=1, n_s=2 \\
[(L_1^{-1}V^TD(m_E+m_T))^T (L_1^{-1}V^TD(m_E+m_T))^T]^T, & \mbox{if } n_c=2, n_s=1 \\
[(L_2^{-1}V^TDm_E)^T (L_2^{-1}V^TDm_T)^T]^T, & \mbox{if } n_c=2, n_s=2 \\
\end{cases} \nonumber\\
H & = & \begin{cases}
2 I , & \mbox{if }  n_c=1, n_s=1\\
[I I]^T, & \mbox{if } n_c=1, n_s=2 \\
[I I], & \mbox{if } n_c=2, n_s=1 \\
I, & \mbox{if } n_c=2, n_s=2 \\
\end{cases}\nonumber\\
M_1 & = & \text{diagn}(U^TD,n_c) \nonumber\\
M_2 & = & \text{diagn}(K_{n_c}, n_c) \nonumber\\
M_3 & = & \text{diagn}(J, n_c) \nonumber\\
M_4 & = & \text{diagn}( J L_{n_s}^{-1} J^T , n_s) \nonumber\\
\end{eqnarray}
where $n_s=1$ for the same-speaker hypothesis, $n_s=2$ for the different-speaker hypothesis, $n_c=1$ for the same-channel hypothesis, $n_c=2$ for the different-channel hypothesis, $L_2 = L_D$, $L_1 = L_S$, $K_2 = K_D = U^TDU+I$ and $K_1 = K_S = 2  U^TDU+I$. 

We can now write down the mean and covariance of the inner posterior for each case:

\begin{eqnarray}
\Sigma_{SC,SS}^{-1} & = & 2 U^T D U + I - 4 J L_S^{-1} J^T \nonumber\\
\Sigma_{SC,DS}^{-1} & = & 2 U^T D U + I - 2 J L_D^{-1} J^T \nonumber\\
\Sigma_{DC,SS}^{-1} & = & \text{diagn}(K_D, 2) - [I I]^T J L_S^{-1} J^T [I I] \nonumber\\
\Sigma_{DC,DS}^{-1} & = & \text{diagn}(K_D, 2) -\text{diagn}( J L_D^{-1} J^T , 2) \nonumber\\
\Phi_{SC,SS} & = & \left((\hat m_E+\hat m_T) - 2 J L_S^{-1} (\tilde m_E+\tilde m_T)\right) \nonumber\\
\Phi_{SC,DS} & = & \left((\hat m_E+\hat m_T) - J L_D^{-1}(\tilde m_E+\tilde m_T)\right) \nonumber\\
\Phi_{DC,SS} & = & \begin{bmatrix}
\hat m_E - J L_S^{-1} (\tilde m_E+\tilde m_T)\\
\hat m_T  - J L_S^{-1}(\tilde m_E+\tilde m_T)
\end{bmatrix}\nonumber\\
\Phi_{DC,DS} & = & \begin{bmatrix} 
\hat m_E - J L_D^{-1} \tilde m_E\\
\hat m_T  - J L_D^{-1}\tilde m_T
\end{bmatrix}\nonumber\\
\end{eqnarray}
where
\begin{eqnarray}
\hat m_E & = & U^TD m_E \nonumber\\
\hat m_T & = & U^TD m_T \nonumber\\
\end{eqnarray}

The log of the inner posterior is then given by
\begin{equation}
\log p(X_{H_{*C}}|M,H_{*S}) = \begin{cases}
-\frac{1}{2}R_xk -  Q_{*C,*S} , & \mbox{if } H_{*C}=H_{SC}\\
-R_xk - Q_{*C,*S} , & \mbox{if } H_{*C}=H_{DC}\\
\end{cases}
\end{equation}
where $Q_{*C,*S} = \frac{1}{2}\log|\Sigma_{*C,*S}| + \frac{1}{2} \Phi_{*C,*S}^T\Sigma_{*C,*S}\Phi_{*C,*S}$.

Finally, since the outer posterior is independent of the channel hypothesis, the logarithm of the LR can be written as a sum of terms involving the outer posterior and the inner posteriors
\begin{equation}
\LLR = \LLR_o + \LLR_i
\end{equation}
where
\begin{eqnarray}
\LLR_o & = & \log \left.{\frac{p(Y_{H_{DS}}|X_{H_{*C}},M)}{p(y)p(Y_{H_{SS}}|X_{H_{*C}},M)}} \right|_{0} \nonumber \\
& = & \log |L_D| - \frac{1}{2}\log |L_S| + \frac{1}{2} \tilde m_E^T (L_S^{-1}-L_D^{-1})  \tilde m_E + \frac{1}{2} \tilde m_T^T (L_S^{-1}-L_D^{-1})  \tilde m_T + \tilde m_T^T L_S^{-1}  \tilde m_E \nonumber
\end{eqnarray}
where we use that $\log p(y)|_0 = -\frac{1}{2}R_y k$, and
\begin{eqnarray}
\LLR_i & = & \log \left. \frac{p(x)p(X_{H_{SC}}|M,{H_{SS}})^{-1}P_{S,S}+ p(x)^2p(X_{H_{DC}}|M,{H_{SS}})^{-1}P_{D,S}}
                      {p(x)p(X_{H_{SC}}|M,{H_{DS}})^{-1}P_{S,D}+ p(x)^2p(X_{H_{DC}}|M,{H_{DS}})^{-1}P_{D,D}} \right|_{0} \nonumber \\
& = & \log \sumexp(\log(P_{S,S}) + Q_{SC,SS}\mbox{, }\log(P_{D,S})+ Q_{DC,SS}) - \nonumber\\
& &  \log \sumexp(\log(P_{S,D})+ Q_{SC,DS}\mbox{, }\log(P_{D,D})+ Q_{DC,DS}) \nonumber
\end{eqnarray}
where $\sumexp(x,y) = \exp(x)+\exp(y)$, and were we use that $\log p(x)|_0 = -\frac{1}{2}R_x k$.

\subsection{Scoring for unseen channels}

In this section we consider a scoring case in which there is a single test sample and $n_E$ enrollment samples from C known channels, all different from the test channel. In this case, the LLR can be written as 

\begin{eqnarray}
LR & = & \frac{p(E,T|H_{SS})}{p(E,T|H_{DS})} \\
& = & \left. \frac{p(X, Y_{H_{DS}}|M)}{p(y)p(X, Y_{H_{SS}}|M)}\right|_0
\end{eqnarray}
where
\begin{equation}
X = \{X_E, x_T\}
\end{equation}
and
\begin{equation}
Y_{H_{*C}} = \begin{cases} 
\{y\},             & \mbox{if } H_{*S} = H_{SS} \\
\{y_E, y_T\}, & \mbox{if } H_{*S} = H_{DS}
\end{cases}
\end{equation}
Now, the posterior is given by 
\begin{eqnarray}
p(X, Y_{H_{*S}}|M) = p(Y_{H_{*S}}|X,M) p(X|M,H_{*S})
\end{eqnarray}
where
\begin{equation}
\left. p(Y_{H_{*S}}|X,M)\right|_0  = \begin{cases} 
N(y=0|L_S^{-1}V^TD(f_E+m_T),L_S^{-1}), & \mbox{if } H_S = H_{SS} \\
N(y=0|L_D^{-1}V^TDf_E,L_D^{-1})N(y=0|L^{-1}V^TDm_T,L^{-1}), & \mbox{if } H_S = H_{DS}
\end{cases}
\end{equation}
where $L_D = n_E V^TDV+I$, $L_S = (n_E+1)  V^TDV+I$, $L = V^TDV+I$, and $f_E = \sum m_E$.
Now we define
\begin{eqnarray}
\tilde f_E & = & V^TD f_E \nonumber\\
\tilde m_T & = & V^TD m_T \nonumber\\
k & = & \log(2\pi) \nonumber
\end{eqnarray}
and rewrite the outer posterior as
\begin{equation}
\left. \log p(Y_{H_{*S}}|X,M)\right|_0  = \begin{cases} 
-\frac{1}{2}R_y k + \frac{1}{2}\log |L_S| - \frac{1}{2} (\tilde f_E + \tilde m_T)^T L_S^{-1}  (\tilde f_E + \tilde m_T), & \mbox{if } H_{*S} = H_{SS} \\
-R_yk + \frac{1}{2} \log |L_D|+ \frac{1}{2} \log |L| - \frac{1}{2} \tilde f_E^T L_D^{-1}  \tilde f_E- \frac{1}{2} \tilde m_T^T L^{-1}  \tilde m_T , & \mbox{if } H_{*S} = H_{DS} \\
\end{cases}
\end{equation}
The inner posterior is computed using Equation  (\ref{eq:innerpost}) with
\begin{eqnarray}
\vec G  & = & \begin{bmatrix} 
	g_1 \\
	.. \\
	g_N \\
	m_T
	\end{bmatrix} \\
\bar{\tilde{\vec Y}} & = & \begin{cases}
\begin{bmatrix}
L_S^{-1} V^T D (f_E+m_T) n_1 \\
... \\
L_S^{-1} V^T D (f_E+m_T) n_C \\
L_S^{-1} V^T D (f_E+m_T)
\end{bmatrix}
& \mbox{if } H_{*S} = H_{SS} \\
\begin{bmatrix}
L_D^{-1} V^T D f_E n_1 \\
... \\
L_D^{-1} V^T D f_E n_C \\
L^{-1} V^T D m_T
\end{bmatrix}
& \mbox{if } H_{*S} = H_{DS} 
\end{cases} \\
H & = & \begin{cases}
\begin{bmatrix}
n_1 I & ... & n_C I & I 
\end{bmatrix} & \mbox{if } H_{*S} = H_{SS} \\
\begin{bmatrix}
n_1 I & ... & n_C I & 0 \\
0 & ... & 0 & I
\end{bmatrix} & \mbox{if } H_{*S} = H_{DS} \\
\end{cases} \\
M_1 & = & \text{diagn}(U^TD,C+1) \\
M_2 & = & \text{diag}(K_1, \ldots, K_C, K) \\
M_3 & = & \text{diagn}(J, C+1)  \\
M_4 & = & \begin{cases}
\text{diag}( J L_S^{-1} J^T) \label{eq:m4} &  \mbox{if } H_{*S} = H_{SS} \\
\text{diag}( J L_D^{-1} J^T, J L^{-1} J^T) \label{eq:m4} &  \mbox{if } H_{*S} = H_{DS} 
\end{cases} \\
\end{eqnarray}
where $K_i = n_i U^TDU+I$, and $K = U^TDU+I$.

The log of the inner posterior is then given by
\begin{equation}
\log p(X|M,H_{*S}) = -\frac{1}{2}R_xk(C+1) -  Q_{*S}
\end{equation}
where 
\begin{equation}
Q_{*S} = \frac{1}{2}\log|\Sigma_{*S}| + \frac{1}{2} \Phi_{*S}^T\Sigma_{*S}\Phi_{*S}
\end{equation}
where the $\Sigma_{*S}$ and $\Phi_{*S}$ can be computed using Equations (\ref{eq:Sigma}) and (\ref{eq:Phi}), respectively, using the $M_1$, $M_2$, $M_3$, $M_4$, $G$ and $\bar{\tilde{\vec Y}}$ and $H$ defined above for the corresponding hypothesis.

Finally, the logarithm of the LR can be written as a sum of terms involving the outer posterior and the inner posteriors
\begin{equation}
\LLR = \LLR_o + \LLR_i
\end{equation}
where
\begin{eqnarray}
\LLR_o & = & \log \left.{\frac{p(Y_{H_{DS}}|X,M)}{p(y)p(Y_{H_{SS}}|X,M)}} \right|_{0} \nonumber \\
& = &\frac{1}{2} \log |L_D| \frac{1}{2} \log |L| - \frac{1}{2}\log |L_S| + \frac{1}{2} \tilde f_E^T (L_S^{-1}-L_D^{-1})  \tilde f_E + \frac{1}{2} \tilde m_T^T (L_S^{-1}-L^{-1})  \tilde m_T + \tilde m_T^T L_S^{-1}  \tilde f_E \nonumber
\end{eqnarray}
where we use  $\log p(y)|_0 = -\frac{1}{2}R_y k$, and
\begin{eqnarray}
\LLR_i & = & \log \left. \frac{p(X|M,{H_{DS}})}{p(X|M,{H_{SS}})} \right|_{0}  =  Q_{SS}- Q_{DS} \nonumber
\end{eqnarray}

\section{Conclusions}

We have proposed a generalization of PLDA for speaker recognition where channel factors are no longer considered independent across samples. This paper derives the formulae necessary to train the model through the expectation-maximization algorithm and to compute likelihood ratios for scoring for a couple of test scenarios.

The proposed method can be used for any task for which standard PLDA is used, when a discrete nuisance factor is known during training. Examples include multi-language speaker recognition using the language labels as the ``channel'' factor, and language recognition using the channel type (say, microphone type) as channel factor. The identity of the channel does not need to be known during scoring since the likelihood-ratio is computed by marginalizing over it.

Experiments using the proposed method for speaker recognition will be reported on a separate publication.

\bibliographystyle{IEEEtran}

\bibliography{all-short}

\begin{thebibliography}{1}
\providecommand{\url}[1]{#1}
\csname url@samestyle\endcsname
\providecommand{\newblock}{\relax}
\providecommand{\bibinfo}[2]{#2}
\providecommand{\BIBentrySTDinterwordspacing}{\spaceskip=0pt\relax}
\providecommand{\BIBentryALTinterwordstretchfactor}{4}
\providecommand{\BIBentryALTinterwordspacing}{\spaceskip=\fontdimen2\font plus
\BIBentryALTinterwordstretchfactor\fontdimen3\font minus
  \fontdimen4\font\relax}
\providecommand{\BIBforeignlanguage}[2]{{%
\expandafter\ifx\csname l@#1\endcsname\relax
\typeout{** WARNING: IEEEtran.bst: No hyphenation pattern has been}%
\typeout{** loaded for the language `#1'. Using the pattern for}%
\typeout{** the default language instead.}%
\else
\language=\csname l@#1\endcsname
\fi
#2}}
\providecommand{\BIBdecl}{\relax}
\BIBdecl

\bibitem{prince:plda}
S.~Prince, ``Probabilistic linear discriminant analysis for inferences about
  identity,'' in \emph{Proceedings of the International Conference on Computer
  Vision}, 2007.

\bibitem{Dehak11}
N.~Dehak, P.~Kenny, R.~Dehak, P.~Dumouchel, and P.~Ouellet, ``Front-end factor
  analysis for speaker verification,'' \emph{IEEE Trans.\ Audio, Speech, and
  Lang.\ Process.}, vol.~19, no.~4, pp. 788--798, May 2011.

\bibitem{sizov2014}
A.~Sizov, K.~A. Lee, and T.~Kinnunen, ``Unifying probabilistic linear
  discriminant analysis variants in biometric authentication,'' in \emph{Joint
  IAPR International Workshops on Statistical Techniques in Pattern Recognition
  (SPR) and Structural and Syntactic Pattern Recognition (SSPR)}.\hskip 1em
  plus 0.5em minus 0.4em\relax Springer, 2014, pp. 464--475.

\bibitem{em4plda}
N.~Brummer, ``{EM} for probabilistic {LDA},'' Available at https://sites.
  google.com/site/nikobrummer, Tech. Rep., 2010.

\bibitem{besag1989}
J.~Besag, ``A candidate's formula: A curious result in bayesian prediction,''
  \emph{Biometrika}, vol.~76, no.~1, pp. 183--183, 1989.

\bibitem{senoussaoui2011}
M.~Senoussaoui, P.~Kenny, N.~Br{\"u}mmer, E.~De~Villiers, and P.~Dumouchel,
  ``Mixture of {PLDA} models in i-vector space for gender-independent speaker
  recognition.'' in \emph{Interspeech}, 2011, pp. 25--28.

\end{thebibliography}

\end{document}